\documentclass[letterpaper, 10 pt, conference]{ieeeconf}
\IEEEoverridecommandlockouts
\usepackage{cite}
\usepackage{amsmath,amssymb,amsfonts}
\usepackage{algorithmic}
\usepackage{graphicx}
\usepackage{textcomp}
\usepackage{xcolor}
\usepackage{glossaries}
\usepackage[T1]{fontenc}
\usepackage{listings}
\usepackage{hyperref}
\usepackage{cleveref}
\usepackage{multirow}
\usepackage{epigraph}
\usepackage{threeparttable}

\crefname{lstlisting}{listing}{listings}
\Crefname{lstlisting}{Listing}{Listings}
\usepackage{array}
\newenvironment{conditions}
  {\par\vspace{\abovedisplayskip}\noindent\begin{tabular}{>{$}l<{$} @{${}={}$} l}}
  {\end{tabular}\par\vspace{\belowdisplayskip}}

\newacronym{llm}{LLM}{Large Language Model}

\definecolor{ocker}{rgb}{0.72, 0.61, 0.31}

\newcommand{\asimo}{\includegraphics[scale=0.08]{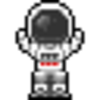}}

\newcolumntype{L}[1]{>{\raggedright\let\newline\\\arraybackslash\hspace{0pt}}m{#1}}

\makeatletter
\def\blfootnote{\gdef\@thefnmark{}\@footnotetext}
\makeatother

\def\BibTeX{{\rm B\kern-.05em{\sc i\kern-.025em b}\kern-.08em
    T\kern-.1667em\lower.7ex\hbox{E}\kern-.125emX}}

\usepackage{fancyhdr}
\fancypagestyle{firstpage}{
	\fancyhf{}
	\fancyfoot[L]{\scriptsize \copyright 2024 IEEE. Personal use of this material is permitted. Permission from IEEE must be obtained for all other uses, in any current or future media, including reprinting/republishing this material for advertising or promotional purposes, creating new collective works, for resale or redistribution to servers or lists, or reuse of any copyrighted component of this work in other works.
		DOI: \href{https://doi.org/10.1109/ICRA57147.2024.10610434}{10.1109/ICRA57147.2024.10610434}}

}

\begin{document}

\title{\Large \bf
CoPAL: \underline{Co}rrective \underline{P}lanning of Robot \underline{A}ctions with \underline{L}arge Language Models\\

\author{Frank Joublin\textsuperscript{\asimo}, Antonello Ceravola\textsuperscript{\asimo},
Pavel Smirnov\textsuperscript{\asimo}, Felix Ocker\textsuperscript{\asimo},
Joerg Deigmoeller\textsuperscript{\asimo},\\Anna Belardinelli\textsuperscript{\asimo},
Chao Wang\textsuperscript{\asimo}, Stephan Hasler\textsuperscript{\asimo},
Daniel Tanneberg\textsuperscript{\asimo}, and Michael Gienger\textsuperscript{\asimo}
\thanks{\asimo~\textit{Honda Research Institute Europe}, Offenbach, Germany
        {\tt\small \{firstname.lastname\}@honda-ri.de}}
        }%
}
\maketitle
\thispagestyle{firstpage}
\begin{abstract}
In the pursuit of fully autonomous robotic systems capable of taking over tasks traditionally performed by humans, the complexity of open-world environments poses a considerable challenge.
Addressing this imperative, this study contributes to the field of Large Language Models (LLMs) applied to task and motion planning for robots.
We propose a system architecture that orchestrates a seamless interplay between multiple cognitive levels, encompassing reasoning, planning, and motion generation.
At its core lies a novel replanning strategy that handles physically grounded, logical, and semantic errors in the generated plans.
We demonstrate the efficacy of the proposed feedback architecture, particularly its impact on executability, correctness, and time complexity via empirical evaluation in the context of a simulation and two intricate real-world scenarios: blocks world, barman and pizza preparation.
\blfootnote{Supplementary material: \url{https://hri-eu.github.io/Loom/}.}
\end{abstract}


\section{Introduction}
Classic planning techniques~\cite{karpas2020automated,brafman2023probabilistic,tanneberg2023learning} focus on searching the optimal task plan, leaving the problems of natural language grounding and low-level motion planning tasks out of scope. 
Integrated task and motion planning (TAMP) approaches~\cite{garret2021integrated} strive to combine high-level reasoning with motion planning for real robots, but might still be challenged by uncertainties and accounting for failure feedback.
Intelligent embodied agents however need to be able to adapt and recover from different kinds of errors.
Large language Models are therefore increasingly used in robotics~\cite{vemprala2023chatgpt,wake2023chatgpt,deepmind2023demonstrating,yoneda2023statler,wu2023tidybot,gramopadhye2023generating,ding2023integrating,driess2023palme,liang2023code, ding2023task} because they provide both very rich commonsense knowledge and implicit reasoning capabilities ~\cite{zhao2023large, ocker2023commonsense, ding2023leveraging}. Advancements and increasingly complex application scenarios demonstrate this impressively~\cite{brown2020language,li2022pre,yang2023harnessing,joublin2023glimpse,yang2023foundation}.
Especially when dealing with underspecified information, LLMs are a promising tool to enable corrective behavior, as they can leverage their inherent world knowledge to provide alternative solutions.
Incorporating feedback to timely reevaluate plans and react to potentially unexpected environmental changes is a crucial requirement for intelligent robots to deal with a wide range of situations.
Leveraging the ability of LLMs to generate different action plan proposals and accounting for previous experience (in the form of prompt updates) appears as a sensible approach for such feedback generation and utilization~\cite{kambhampati2023can}:
    \textit{"The trick is to recognize that LLMs are generating potential answers to be checked/refined by external solvers, .."} 
The present paper makes two contributions to close this gap by proposing a hierarchical architecture for robot manipulation tasks involving success/failure checks:
\begin{itemize}
    \item A novel closed-loop task planning mechanism with a multi-level feedback loop (CoPAL). 
    \item An evaluation of the planning mechanism demonstrating how different kinds of low-level feedback improve the quality of planning and execution in three different scenarios, both in simulation and on the real robot.
\end{itemize}

\begin{figure}[t]  
    \centering
    \includegraphics[width=\columnwidth]{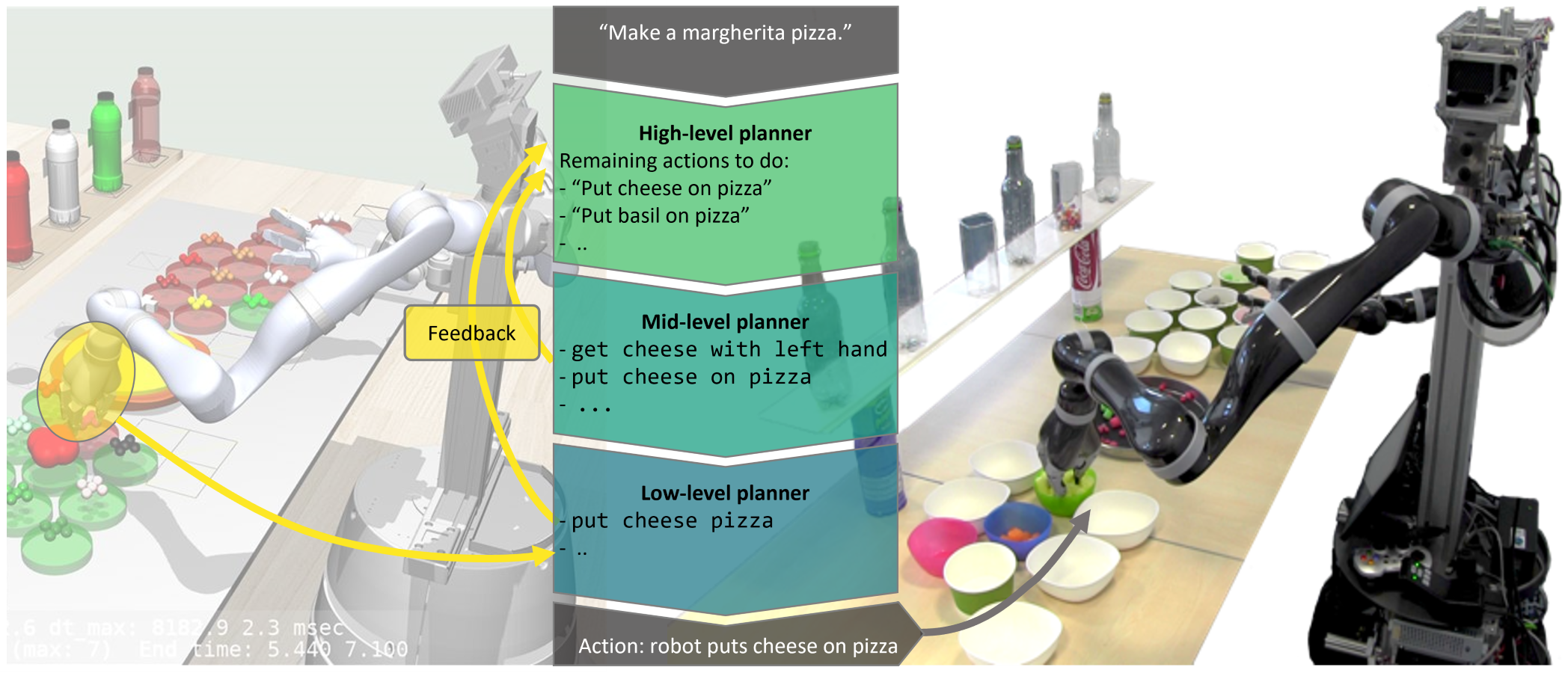} 
    \caption{Snapshot of a pizza domain task execution on the real robot alongside an illustrative description of the proposed framework.}
    \label{fig:robot_framework}
\end{figure}

\section{Related work}\label{sec:related_work}
The idea of using LLMs for grounding high-level user goals into mid-level action sequences (skills or action primitives available to a robot) has been recently investigated in~\cite{huang2022language, xie2023translating, singh2023progprompt,wang2023describe }. Plain planning capabilities of LLMs~\cite{valmeekam2022large, liu2023llm+}, backprompting~\cite{valmeekam2023planning, guan2023leveraging }, and corrective reprompting strategies~\cite{raman2022planning} have been studied as well. While these studies demonstrate how to efficiently utilize LLMs for planning, they do not offer a closed-loop planning architecture on all levels, where geometric configuration of an environment as well as task execution feedbacks are taken into account.
Here, we aim to address exactly this gap. Below we review a number of studies targeting closed-loop planning mechanisms, with advances in feedback handling. 

Ichter et al. presented SayCan~\cite{ichter2023} - an approach for grounding LLMs in affordance functions, by capturing probabilities of possible skills to be useful for high-level goal achievement and to be executed successfully from the current state. The approach relies on training skills and their policies via reinforcement learning, which could be considered a costly step in the era of large language models like GPT-4. The authors state that SayCan receives environmental feedback only at a current decision step but in case a skill fails or the environment unexpectedly changes, the feedback is not available. In this sense, feedback obtained from our simulation system during the planning loop promptly triggers corrective actions handling failing skills during planning.

Huang et al. put forward InnerMonologue~\cite{huang2022inner} - a study, where authors investigate to which extent an LLM planner is able to utilize environment feedback, such as success/failure, object detection, scene description, visual question answering, and human feedback. Their approach is comparable to ours in terms of incorporating various types of feedback into the planning loop, however, that framework does not incorporate low-level geometric constraints and feedback from a motion planning system, which both are critical aspects for planning on a real robot

Lin et al. pointed out a limitation of SayCan and InnerMonologue in terms of generating optimal long-horizon plans~\cite{lin2023text2motion}. There, the Text2Motion approach is proposed, which utilizes LLMs for predicting symbolic goal states and then applies one of two planning strategies. The first one prompts LLMs for a plan considered to be optimal according to a certain objective. Each skill (mid-level task)  in the returned plan should pass through geometric feasibility checks (probability of being executed from the current state) otherwise it gets excluded from the plan. If there are no skills left and the plan is empty, a greedy iterative strategy (similar to the one in SayCan) is applied as a fallback solution. In contrast to Text2Motion, our approach handles non-feasible skills instead of excluding them from the plan. As a geometric feasibility checker, we use a simulator, which provides feedback about failing steps in the plan. More details are provided in \Cref{sec:planning_concept}. 

Close to our approach is SayPlan, proposed by Rana et. al.~\cite{rana2023sayplan}. It relies on a search over a 3D scene graph, provisions LLMs with found subgraphs, and starts an iterative replanning based on feedback generated by the scene graph simulator. In comparison to our study, the authors focus on semantic search and reasoning, which is performed outside of the LLM, leaving details about feedback out of focus. 

In summary, most related work focuses on solving the challenges of symbolic planning, while only a few work has worked towards propagating the plans down to the physical world.
To the best of our knowledge, there is no in-depth analysis of how the resulting low-level feedback can be leveraged for resolving issues. Here, we tackle this gap by proposing a hierarchical replanning architecture for long-horizon robot manipulation tasks that gathers feedback from all levels down to the motion planning.

\section{Corrective Planning Concept}\label{sec:planning_concept}
\begin{figure*}[t]  
    \centering
    \includegraphics[width=\textwidth]{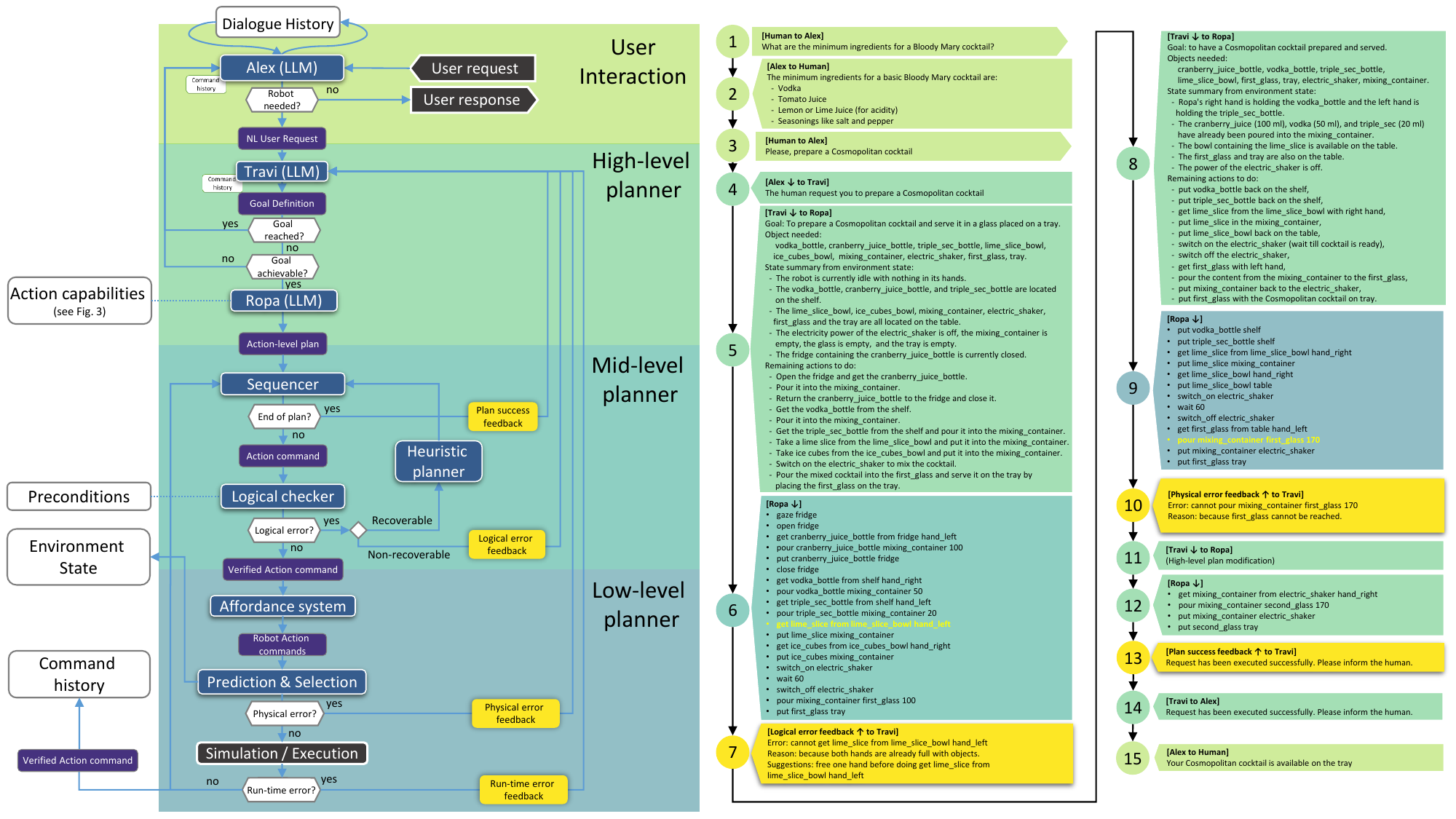} 
    \caption{Left: System architecture. Computational modules are drawn in blue, top-down instructions in green, bottom-up feedback in orange, and real-world interaction modules in dark green. Right: Execution flow for two exemplary human requests (step 1 and 3). }
    \label{fig:flow}
\end{figure*}

The proposed system, called CoPAL, is tailored to the information types available in robot planning, specifically natural language instructions and semantic, symbolic, and sensory feedback.
We found that a decomposition of the planning problem into four respective dedicated layers (see \Cref{fig:flow}) has significant advantages.
This decomposition enables the definition of clear roles for the individual modules, which turns out to be particularly beneficial for the LLM-based agents.
These agents can be instantiated with clear system messages, resulting in sound encapsulation.
The benefit of encapsulation is not only a software engineering paradigm, but is also supported for agents by related work~\cite{patra2021deliberative} and our own experiments.
The system's first and second layers include three LLM-based conversational agents, leveraging world knowledge to transform incomplete and unspecified human instructions into a long-horizon action plan.
The third and fourth layer are responsible for scheduling, prediction, decision-making, and grounding the action plan in the physical capabilities of the robot and in the 3D environment. 
The lowest layer updates the environment state and makes it available to all upper levels.
It contains a geometric representation of the scene, including coordinates of the robot, available objects, and other relevant information such as parent-child hierarchies.
The layers are described in more detail in the following.

\subsection{User Interaction Layer}
The User Interaction Layer acts as the mediator between the human user and the rest of the system. The first LLM agent, which we call \texttt{Alex}, is prompted to be a natural language interpreter of human request, and to provide a direct answer for requests that do not require any physical response. Otherwise, to forwards the request to the next LLM agent. 
\texttt{Alex} is the only agent that makes use of dialogue history, which allows it to communicate robot feedback in relation to the original human requests. It also maintains a history of the performed actions in order to enrich its response to the human.

\subsection{High-Level Planner}
The role of the second LLM agent, called \texttt{Travi}, is to be responsible for translating the request from \texttt{Alex} into a high-level plan with detailed specifications for the next LLM agent. It is prompted to generate in a chain-of-thought (CoT) way a request with the following information: the goal of the human request, the list of objects needed for achieving this goal (extracted from the environment state), a summary of the current environment state, and the remaining steps to be executed. These consecutive steps form a broader plan that is formulated in a descriptive language. Forcing the LLM to output all this information helps it to generate a meaningful high level plan. Another function of \texttt{Travi} is to integrate feedback from the lower layers into the planning phase and adjust plans as necessary or even replan. Delivering natural language feedback to the LLM regarding environmental changes is termed backprompting, as discussed in \Cref{sec:related_work}. The implementation of backprompting has demonstrated notable enhancements in planning challenges, such as the blocks world problem, which we further explore in \Cref{sec:blocksworld}.

The third LLM agent, called \texttt{Ropa}, receives \texttt{Travi's} high-level plan specification and generates a high-level plan from it. This involves translating overarching goals into actions, which the lower layers can process. For the investigated scenarios, we used the action commands specified in \Cref{tab:api}. In case no plan can be generated (i.e., the goal state is already achieved or unachievable by the robot), a feedback message is sent to \texttt{Alex} to inform the human.
The amount of replanning cycles is limited to a fixed number (5 in our experiments) of attempts before aborting the planning process and informing the human.

\begin{table}[tb]
\caption{Robot Action Capabilities*}
\label{tab:api}
\scriptsize
\begin{threeparttable}
\begin{tabular}{L{.43\columnwidth} L{.47\columnwidth}}
\textbf{Command} & \textbf{Description} \\
\hline
get <object1> from <object2> (<hand>) & take object1 from object2 using hand \\
\hline
put <object1> <object2> & put object1 (in hand) on/in object2 \\
\hline
pour <object1> <object2> <x [ml]> & pour x milliliter (liquid/solid) from object1 (in hand) into object2 \\
\hline
open\_door <object1> & open door of object1 (microwave) \\
\hline
close\_door <object1> & close door of object1 (microwave) \\
\hline
screw <object1> & close object1 (bottle) by screwing tap \\
\hline
unscrew <object1> & open object1 (bottle) by unscrewing tap \\
\hline
finger\_push <object1> & push object1 with fingertip (to power on/off devices) \\
\hline
wait <x [s]> & wait for x seconds
\end{tabular}
\end{threeparttable}
\begin{tablenotes}\footnotesize
\item[*] * Action commands: the action-level plan (Fig.~\ref{fig:flow} between high- and mid-level planner) consists of a sequence of these. The individual actions are executed by the robot via the low-level planner.
\end{tablenotes}
\end{table}

\subsection{Mid-Level Planner}
The mid-level planner is required to compensate potentially incomplete, incorrect, or redundant sequences of commands coming from the higher layer. It sequentially processes each action of the plan one after the other (therefore it is named Sequencer). A Logical Checker verifies each action command in terms of syntax, object names, and preconditions. Two types of errors are considered: recoverable (see \Cref{tab:recoverable}) and unrecoverable errors (see \Cref{tab:unrecoverable}). Recoverable errors are handled by the Heuristic Planner by modifying the mid-level plan using insertion or substitution of commands. A rule-based algorithm is currently being used for this step.  Unrecoverable errors lead to a replanning in the high-level planning layer. 

\begin{table}[b]
\caption{Examples of Recoverable Errors}
\scriptsize
\begin{center}
\begin{tabular}{L{.2\columnwidth} L{.2\columnwidth} L{.4\columnwidth}}
\textbf{Command} & \textbf{Failure} & \textbf{Mid-Level Plan Modifications} \\
\hline
put object1 object2 & object1 is not in hand & free one hand if needed, then take object1, put object1 object2 and possibly get previous object \\
\hline
put object1 object2 & object2 is in closed container & free one hand if needed, then open container, put object1 object2, close container and possibly get previous object \\
\hline
open\_door object1 & all hands are full  & free one hand, open\_door object1 and possibly get previous object \\
\end{tabular}
\label{tab:recoverable}
\end{center}
\end{table}

\begin{table}[tb]
\caption{Examples of Unrecoverable Errors}
\scriptsize
\begin{center}
\begin{tabular}{L{.35\columnwidth} L{.5\columnwidth}}
\textbf{Type} & \textbf{Examples of errors} \\
\hline
Syntactic/ Semantic Errors & Wrong command syntax \newline Wrong object names \\
\hline
Logical Errors & Getting object with all hands full \newline Pouring into an object placed in a container \newline Pouring into a full object \\
\hline
Physical Errors & Joint limits errors \newline Obstacle object on the way \newline Object out of reach \\
\end{tabular}
\label{tab:unrecoverable}
\end{center}
\end{table}

\subsection{Low-Level Planner}
The role of the lowest layer of this planning hierarchy is to bridge the gap between the textual action instruction as shown in \Cref{tab:api}, and its grounding in the motion generation that considers the 3D scene, the physical properties of the objects as well as the manipulation capabilities of the robot. It comprises an affordance system that instantiates the commands as a set of potential primitives that achieve the action goal. For example, the action "get bottle" can be instantiated for a bi-manual robot by four primitives: Getting the bottle with the left or right hand, each with a top-grasp or power grasp. Each primitive is evaluated in a simulation with respect to its feasibility and ranked according to a collision and joint limit metric~\cite{gienger2010whole}. The winning primitive is selected for execution in simulation. In case all primitives fail, the reason for failure of the best primitive is communicated. Feedback on this level is related to physical errors and run-time errors. Physical errors comprise issues such as unreachable objects, collisions between robot and the environment, violations of robot limits, and so on. Run-time errors relate for instance to hardware failures or other unmodelled issues, and only plays a role if the system is operated with a real robot. It is assumed that the errors generated on this level cannot easily be recovered. Therefore they are communicated to the high-level planner, where they typically lead to a plan change.

\subsection{Feedback Loops}\label{sec:feedback}
When the user asks an epistemic question (e.g. Step 1 of \Cref{fig:flow}) \texttt{Alex} provides the answer in natural language similarly to a chatbot.  We consider this an \textit{epistemic feedback}, which we do not further evaluate. On the other hand, as soon as the system is asked to carry out a physical task (e.g., preparing a cocktail, Step 3), a series of planning agents (\texttt{Travi}, \texttt{Ropa}) are triggered in sequence, until a first error is generated, which can happen at different levels, as seen in the previous subsections. While mid-level recoverable errors are handled internally by the heuristic planner (yielding a mid-level plan revision), unrecoverable mid-level errors and physical errors call for  a high-level replanning, hence feedback messages are sent to \texttt{Travi} for reprompting. \textit{Logic error feedback}, \textit{physical error feedback}, and \textit{run-time error feedback} are formulated as triplet $\langle E,R,S\rangle$, in a string: ``Error:  $\langle$what command failed$\rangle$, Reason: $\langle$why it failed$\rangle$, Suggestion: $\langle$how to resolve the problem$\rangle$.''.

\section{Experiments}

To evaluate the proposed architecture, we conducted several experiments using OpenAI's GPT-4 via the API.
We extended the barman planning problem~\cite{liu2023llm+}, conducted experiments for preparing pizzas, and experimented with the blocks world problem.

\subsection{Barman Experiments}
In this scenario, the LLM-backed robot identifies necessary ingredients, comes up with a plan, executes the plan accordingly, and replans as necessary.
For evaluation purposes, we selected recipes as ground truth plans for the barman and the pizza experiments.
We identified ten cocktails of interest: Bloody Mary, Caipirinha, Cosmopolitan, Daiquiri, Gin and Tonic, Long Island Iced Tea, Manhattan, Margarita, Martini, and Mojito.
For these, we specified the ingredients, distinguishing between liquid, solid, and optional ingredients.
The complete list entails: bitters, coke, cranberry juice, gin, ice cubes, lemon slice, lime slice, mint, orange slice, rum, salt, syrup, sliced lemon, soda water, sugar cubes, tequila, tomato juice, tonic water, triple sec, vermouth, vodka, and whiskey.
For instance, \Cref{lst:negroni} shows the expected ingredients for a Cosmopolitan.

\begin{lstlisting}[caption={Ground truth ingredients for a Cosmopolitan.}, label={lst:negroni}, captionpos=b, basicstyle=\scriptsize]
"02.03 Cosmopolitan": {
    "ingredient": ["lime_slice"],
    "liquid": ["cranberry_juice", "triple_sec", "vodka"],
    "optional": ["ice_cubes"]
},
\end{lstlisting}

For assessing the impact of the types of feedback and the different replanning levels, we considered several system configurations, as defined in  \Cref{tab:ablation}.

\begin{table}[tb]
\caption{System configurations}
\scriptsize
\begin{center}
\begin{tabular}{l p{.75\columnwidth}}
\textbf{Symbol} & \textbf{Description} \\
\hline
BL & Baseline, neither mid-level nor high-level replanning \\
\hline
M & Only mid-level planner for replanning \\
\hline
H\textsubscript{0} & Only high-level planner for replanning \newline Feedback: what (cf. \Cref{sec:feedback})  \\
\hline
H\textsubscript{1} & Only high-level planner for replanning \newline Feedback: what + why (cf. \Cref{sec:feedback}) \\
\hline
H\textsubscript{2} & Only high-level planner for replanning \newline Feedback: what + why + how (cf. \Cref{sec:feedback}) \\
\hline
MH\textsubscript{i} & Combination of M and H\textsubscript{i} \\
\end{tabular}
\label{tab:ablation}
\end{center}
\end{table}

To allow for creative planning in the kitchen setup, we used OpenAI's default temperature of $0.8$.
For each cocktail and every configuration involving replanning via the high-level planner, we conducted 10 runs, resulting in 600 trials in total.
We compute the results for the baseline (BL) and replanning using only the mid-level planner (M) from the available data, by considering every plan requiring high-level replanning as not executable.

\begin{figure}[tbp]
\centerline{\includegraphics[width=\columnwidth]{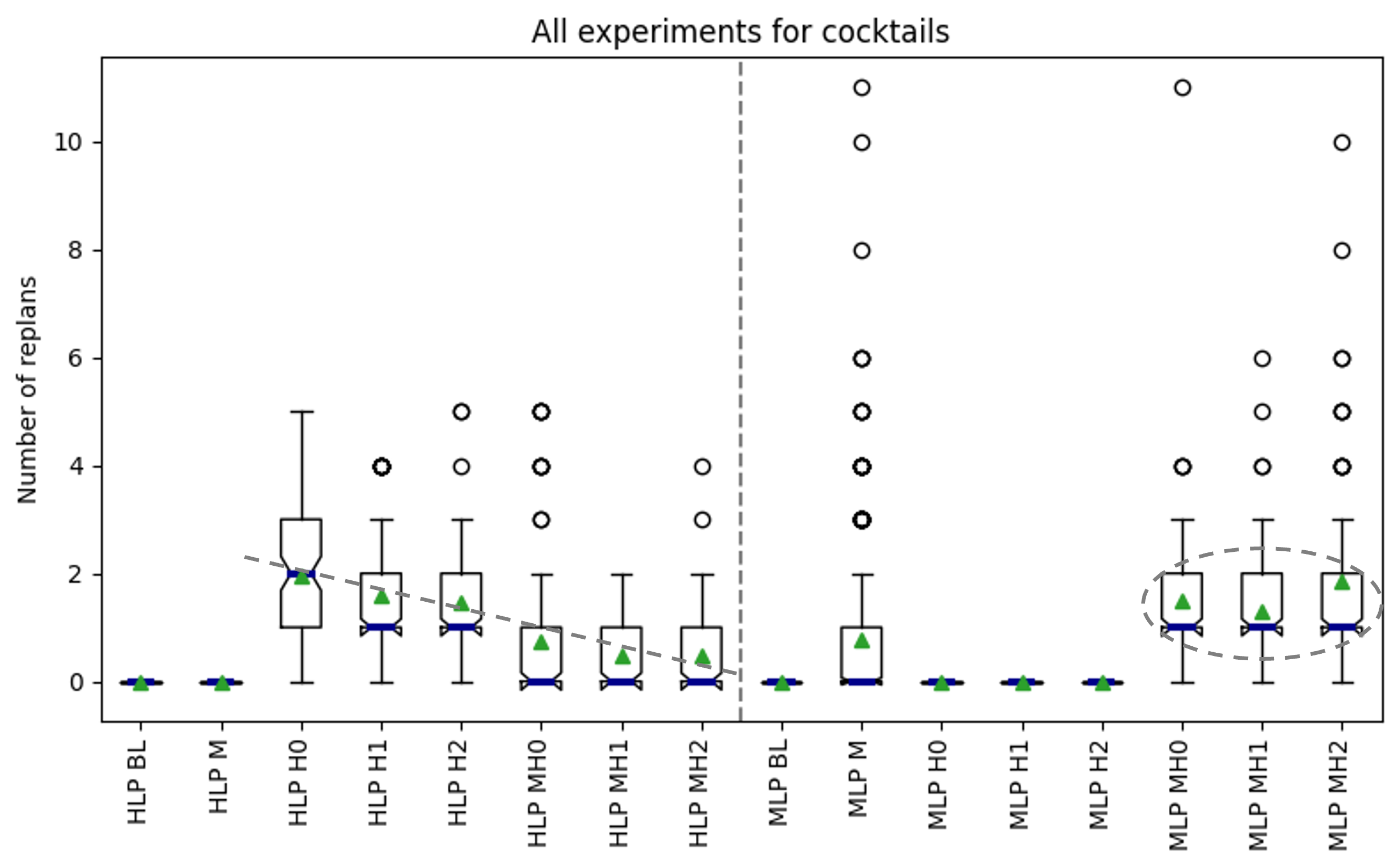}}
\caption{Amount of high-level (HLP, left) and mid-level (MLP, right) replanning by setup.}
\label{fig:replansubplan}
\end{figure}

\Cref{fig:replansubplan} shows the number of replans for the various setups, differentiating between high-level and mid-level replanning.
Per definition, there is no replanning whatsoever in the baseline, and no replanning using the mid-level planner for the H\textsubscript{i} variations.
It becomes apparent that improved feedback quality reduces the amount of necessary high-level replanning, see the trend highlighted by the dashed line on the left of \Cref{fig:replansubplan}.
Also, using mid-level replanning for recoverable errors (MLP for MH\textsubscript{i}) reduces the amount of necessary high-level replanning (HLP for MH\textsubscript{i}).

\begin{figure}[tbp]
\centerline{\includegraphics[width=\columnwidth]{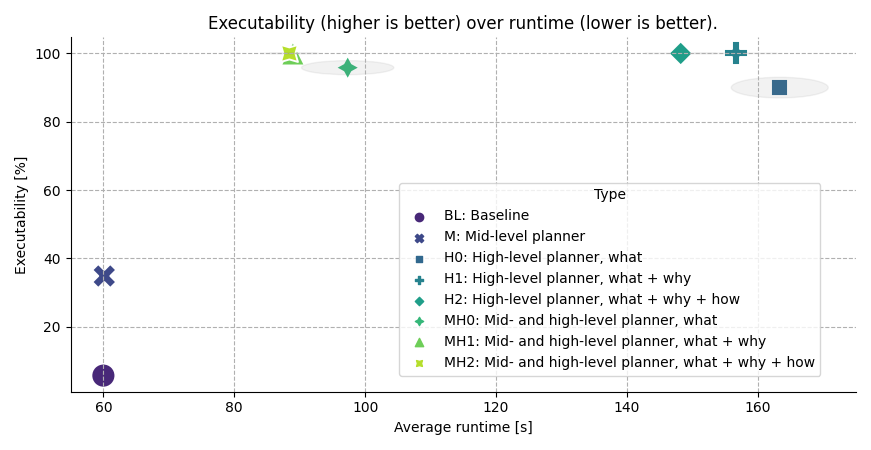}}
\caption{Comparison of the architecture variations regarding executability and runtimes: architectures replanning both on a mid- and high-level achieve the best trade-off between runtime and executability.}
\label{fig:executability-and-times}
\end{figure}

We evaluate the performance of the architecture variations by considering plan executability, and the runtimes for generating the plans, cf. \Cref{fig:executability-and-times}.
Here, a plan is considered executable, if the robot does not end up in a loop exceeding five replanning steps (done by high-level planner).
Without any replanning (BL), the LLM achieves only 5.7~\% executability.
While replanning with the mid-level planner (M) increases executability to 34.8~\%, using a high-level planner, e.g., MH\textsubscript{1}, always yields executable plans.
Note that more detailed feedback also positively influences executability (MH\textsubscript{1} versus MH\textsubscript{0}), and the resolution of recoverable errors via the mid-level planner increases executability even if the feedback quality is limited (MH\textsubscript{0} versus H\textsubscript{0}).

Considering just the runtime metric, the analysis is strongly affected by the required amount of planning (Duration of GPT3.5 for \texttt{Alex}: $0.6\pm0.1s$, of GPT4 for \texttt{Travi}: $16.1\pm5.1s$, and for \texttt{Ropa}: $14.9\pm3.3s$): generating a plan dominates the time complexity over the heuristic mid-level planner by about 1-2 orders of magnitude.
As indicated in \Cref{fig:executability-and-times}, the baseline and the mid-level planner are particularly fast because they rely on prompting the LLMs only once to generate a plan.
Critically, leveraging the efficient mid-level planner in combination with the high-level planner (MH\textsubscript{i}) is significantly quicker than using only the high-level planner (H\textsubscript{i}).
Note that the quality of the feedback also has a positive effect on the runtimes (MH\textsubscript{2} versus MH\textsubscript{0}).

We consider plan correctness as another performance metric.
Correctness in the barman scenario is defined as following the given plan, i.e., placing exactly one glass on the tray, and including all mandatory ingredients but nothing that is not an optional ingredients.
The average correctness for processes that were executable was 87~\% across all architecture types, indicating that correctness without final feedback regarding satisfaction from the human is independent from executability.
Overall, the proposed architecture performed very well, considering that the user does not specify a recipe, but the LLM draws from its world knowledge to identify relevant ingredients.

We conducted an in-depth analysis of incorrect results using a distance metric based on the number of incorrect ingredients, cf. \Cref{eq:editdistance}.
Missing ingredients, e.g., a Cosmopolitan without triple sec, are considered worse than superfluous ingredients, e.g., a Bloody Mary with basil, and are weighted respectively ($1.0$ versus $.2$).
This analysis shows that 5~\% of incorrect cocktails are caused by missing ingredients ($d=1$) and 5~\% are caused by superfluous ingredients that were not optionally allowed ($d=.2$).
The number of multiple errors is significantly lower, with about 1~\% each for two missing ingredients, and one missing and one superfluous ingredient.

\begin{equation}
d = \sum_{e \in E} w_{s} \cdot s_{e} + w_{m} \cdot m_{e}
\label{eq:editdistance}
\end{equation}
where:
\begin{conditions}
E & experiments \\
w_{s} & weighting factor for superfluous ingredients ($.2$) \\
s & number of superfluous ingredients \\
w_{m} & weighting factor for missing ingredients ($1.0$) \\
m & number of missing ingredients
\end{conditions}

\subsection{Blocks world Experiment}\label{sec:blocksworld}

In the second experiment, we utilized a dataset from~\cite{liu2023llm+} for an advanced block-stacking experiment, which is characterized by a given goal configuration of blocks. This well-established blocks world problem poses challenges due to the required order of blocks in the goal configuration. 

This experiment shows the generalization capabilities of the presented approach. We do not incorporate any model of pre-conditions as commonly done by classical problem solvers operating with PDDL~\cite{mcdermott_pddl_1998,haslum2019introduction}. Instead, our approach relies on the low-level layer of our system, which makes it transferable to other problem domains. Placement constraints are considered by the employed affordance model. Invalid configurations like stacking on an occupied place are identified by collision detection and fed back as physical errors to the high-level planner. The state of the blocks configuration is translated from the environment state to natural language to assess if the desired goal configuration has been achieved. This step is carried out in the high-level planning layer. 

\Cref{tab:blocks world} summarizes the results of our architecture on the blocks world experiment including three blocks (left column) and up to 5 blocks (3, 4 and 5 blocks). 
We reached a similar performance of $82\%$ for less or equal to 4 blocks as reported in~\cite{valmeekam2023planning}. This marks a significant enhancement when compared to~\cite{liu2023llm+}, probably because we use CoT, back-prompting and more elaborate task description in our prompts. On the other hand, the performance drops significantly for 5 blocks. This indicates that on a short horizon logical problem, the combination of LLMs with detailed and tightly, step-wise incorporated feedback leads to very good results. For complex planning problems, like 5 blocks and more, we empirically observed that LLMs are not good at solving the task. It seems to us that LLM planning are good for non-purely algorithmic tasks where common sense knowledge can be used as shown in the next experiment. In this case the planning horizon can be long.

\begin{table}[tb]
\caption{Success Rates on blocks world Experiment}
\scriptsize
\begin{center}
\begin{tabular}{c | c | L{.5\columnwidth}}
\multicolumn{2}{c|}{\textbf{Success rate}} & \multirow{2}{*}{\parbox{.5\columnwidth}{\textbf{Description of the method}}} \\
\textbf{3 blocks} & \textbf{5 blocks} \\
\hline
\multicolumn{2}{c | }{\textit{CoPAL (Ours)}} & \multirow{2}{*}{\parbox{.5\columnwidth}{Backprompt after fully executed plan and evaluate goal using GPT-4}} \\
83\% & 73\% \\
\hline
\multicolumn{2}{c | }{\textit{LLM+P}} & \multirow{2}{*}{\parbox{.5\columnwidth}{Classical planner combined with GPT-3 \cite{liu2023llm+}}} \\
33\% & 46\% & \\
\hline
\multicolumn{2}{c | }{\textit{LLM-As-P}} & \multirow{2}{*}{\parbox{.5\columnwidth}{Plain GPT-3 as planner \cite{liu2023llm+}}} \\
0\% & 0\% & \\
\end{tabular}
\label{tab:blocks world}
\end{center}
\end{table}

\subsection{Robot-in-the-loop Experiments}

\begin{figure}[tbp]
\centerline{\includegraphics[width=\columnwidth]{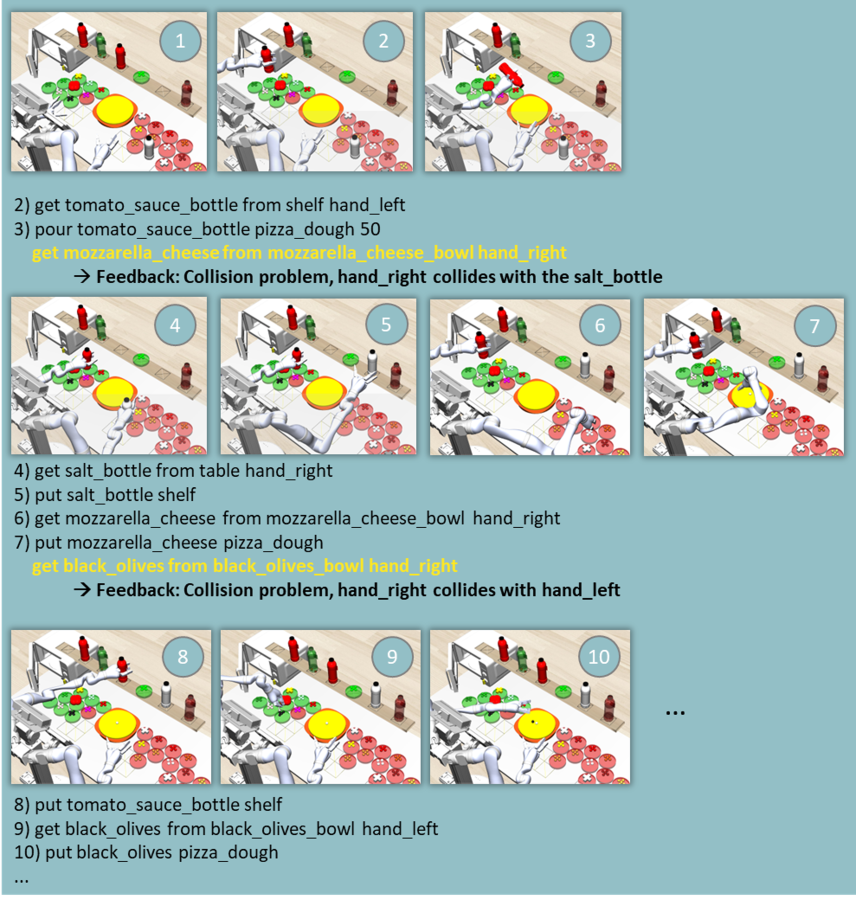}}
\caption{Extract of an experiment with collisions and low-level feedback.}
\label{fig:low-level feedback}
\end{figure}

In line with the previous blocks world experiment, we created a ground truth for the pizza scenario with respective ingredients, such as dough, tomato sauce, and mushrooms, analogously to the barman scenario and run several experiments. 
The scenario was configured such that larger objects were blocking the reachability of food items, requiring a re-planning to remove obstacles.
\Cref{fig:low-level feedback} shows part of an experiment where two collisions are detected by the low-level planner and corresponding replan provided by the LLM. 

Figure~\ref{fig:low-level feedback} illustrates the low-level feedback and its interplay with the high-level planning layer, finally resolving the situation by putting the occluding salt on the shelf (steps 4 and 5), and putting away the tomato sauce bottle (step 8) to utilize the left hand to pick up the black olives. We find this a quite remarkable emergent problem-solving behavior. 

The experiments have also been executed with a real robot (see \Cref{fig:robot_framework}). We used soft material for the ingredients used by the real robot. The full architecture easily runs on a labtop. Several videos of running experiments are available in supplementary materials.

\section{Summary and Outlook}

At the heart of this paper is a novel LLM-based replanning strategy, that leverages multi-level feedback with the physical robot in a closed loop.
We demonstrated the performance of the planning mechanism in three different scenarios, both in simulation and on the real robot.
First, high-level replanning using an LLM significantly increases executability.
Second, the quality of the feedback, and specifically low-level feedback from the sensory level, improves the executability of the generated plans and also the execution times.
Third, runtimes are positively influenced by introducing a quick, heuristic-based mid-level planning that resolves traditional action precondition errors.
Finally, this paper demonstrates the potential of LLM-based robots to act in an open world, reacting appropriately to unforeseen issues.

Extending these findings, future work should include research regarding explainability, e.g., via the combination of LLM-based approaches for intuitive reactions and GOFAI for optimal planning.

As we probe deeper into LLMs, prompt design emerges as critical. Our methodology minimized domain-centric biases and sharply defined the role of each LLM agent (see https://hri-eu.github.io/Loom/ for a description of all prompts). Back-prompting plays a significant role and although the performance improvement given by the suggestion part in the feedback is little, we notice a qualitative sensitivity to feedback coming specifically from the low-level planner where the LLMs have to become more creative depending on the situation (obstacle, far-reaching-objects, tools requirements,...). 

Future directions may encompass the augmentation of our system to include mechanisms adept at extracting examples prompts, as hinted by research such as \cite{bharadhwaj2023roboagent}.

Notwithstanding the achievements, we noted that latencies associated with the GPT-4 API are still a concern. Although the inference time of OpenAI API is steadily reducing, a way to deal with it in Robotic application now is to introduce speech and behavior filler that makes the robot behave or communicate during API queries. Nevertheless, initiating replanning during ongoing valid steps, as observed in the mid-level planner, reduces latency-induced delays. To further alleviate LLM-associated latencies, a promising avenue would be the amplification of memory usage and learning from past planning scenarios.

Finally, the system should be extended for safe cooperation with humans in the sense of cooperative intelligence~\cite{sendhoff2020cooperative}.


\addtolength{\textheight}{-0cm}  

\bibliographystyle{IEEEtran}
\bibliography{root.bib}

\begin{thebibliography}{10}
\providecommand{\url}[1]{#1}
\csname url@samestyle\endcsname
\providecommand{\newblock}{\relax}
\providecommand{\bibinfo}[2]{#2}
\providecommand{\BIBentrySTDinterwordspacing}{\spaceskip=0pt\relax}
\providecommand{\BIBentryALTinterwordstretchfactor}{4}
\providecommand{\BIBentryALTinterwordspacing}{\spaceskip=\fontdimen2\font plus
\BIBentryALTinterwordstretchfactor\fontdimen3\font minus
  \fontdimen4\font\relax}
\providecommand{\BIBforeignlanguage}[2]{{%
\expandafter\ifx\csname l@#1\endcsname\relax
\typeout{** WARNING: IEEEtran.bst: No hyphenation pattern has been}%
\typeout{** loaded for the language `#1'. Using the pattern for}%
\typeout{** the default language instead.}%
\else
\language=\csname l@#1\endcsname
\fi
#2}}
\providecommand{\BIBdecl}{\relax}
\BIBdecl

\bibitem{karpas2020automated}
E.~Karpas and D.~Magazzeni, ``Automated planning for robotics,'' \emph{Annual
  Review of Control, Robotics, and Autonomous Systems}, 2020.

\bibitem{brafman2023probabilistic}
R.~I. Brafman, D.~Tolpin, and O.~Wertheim, ``Probabilistic programs as an
  action description language,'' \emph{AAAI Conference on Artificial
  Intelligence}, 2023.

\bibitem{tanneberg2023learning}
D.~Tanneberg and M.~Gienger, ``Learning type-generalized actions for symbolic
  planning,'' \emph{IEEE/RSJ International Conference on Intelligent Robots and
  Systems (IROS)}, 2023.

\bibitem{garret2021integrated}
C.~R. Garrett, R.~Chitnis, R.~Holladay, B.~Kim, T.~Silver, L.~P. Kaelbling, and
  T.~Lozano-P\'{e}rez, ``Integrated task and motion planning,'' \emph{Annual
  Review of Control, Robotics, and Autonomous Systems}, 2021.

\bibitem{vemprala2023chatgpt}
S.~Vemprala, R.~Bonatti, A.~Bucker, and A.~Kapoor, ``{ChatGPT} for robotics:
  Design principles and model abilities,'' Microsoft, Tech. Rep., 2023.

\bibitem{wake2023chatgpt}
N.~Wake, A.~Kanehira, K.~Sasabuchi, J.~Takamatsu, and K.~Ikeuchi, ``{ChatGPT}
  empowered long-step robot control in various environments: A case
  application,'' \emph{IEEE Access}, 2023.

\bibitem{deepmind2023demonstrating}
G.~DeepMind, ``Demonstrating {Large Language Models} on robots,''
  \emph{Robotics: Science and Systems (RSS) Demo Track}, 2023.

\bibitem{yoneda2023statler}
T.~Yoneda, J.~Fang, P.~Li, H.~Zhang, T.~Jiang, S.~Lin, B.~Picker, D.~Yunis,
  H.~Mei, and M.~R. Walter, ``Statler: State-maintaining language models for
  embodied reasoning,'' \emph{arXiv preprint arXiv:2306.17840}, 2023.

\bibitem{wu2023tidybot}
J.~Wu, R.~Antonova, A.~Kan, M.~Lepert, A.~Zeng, S.~Song, J.~Bohg,
  S.~Rusinkiewicz, and T.~Funkhouser, ``Tidybot: Personalized robot assistance
  with large language models,'' \emph{IEEE/RSJ International Conference on
  Intelligent Robots and Systems (IROS)}, 2023.

\bibitem{gramopadhye2023generating}
M.~Gramopadhye and D.~Szafir, ``Generating executable action plans with
  environmentally-aware language models,'' \emph{arXiv preprint
  arXiv:2210.04964}, 2023.

\bibitem{ding2023integrating}
Y.~Ding, X.~Zhang, S.~Amiri, N.~Cao, H.~Yang, A.~Kaminski, C.~Esselink, and
  S.~Zhang, ``Integrating action knowledge and {LLMs} for task planning and
  situation handling in open worlds,'' \emph{Autonomous Robots (2023)}, 2023.

\bibitem{driess2023palme}
D.~Driess, F.~Xia, M.~S.~M. Sajjadi, C.~Lynch, A.~Chowdhery, B.~Ichter,
  A.~Wahid, J.~Tompson, Q.~Vuong, T.~Yu, W.~Huang, Y.~Chebotar, P.~Sermanet,
  D.~Duckworth, S.~Levine, V.~Vanhoucke, K.~Hausman, M.~Toussaint, K.~Greff,
  A.~Zeng, I.~Mordatch, and P.~Florence, ``Palm-e: An embodied multimodal
  language model,'' \emph{International Conference on Machine Learning}, 2023.

\bibitem{liang2023code}
J.~Liang, W.~Huang, F.~Xia, P.~Xu, K.~Hausman, B.~Ichter, P.~Florence, and
  A.~Zeng, ``Code as policies: Language model programs for embodied control,''
  \emph{IEEE International Conference on Robotics and Automation (ICRA)}, 2023.

\bibitem{ding2023task}
Y.~Ding, X.~Zhang, C.~Paxton, and S.~Zhang, ``Task and motion planning with
  large language models for object rearrangement,'' \emph{arXiv preprint
  arXiv:2303.06247}, 2023.

\bibitem{zhao2023large}
Z.~Zhao, W.~S. Lee, and D.~Hsu, ``{Large Language Models} as commonsense
  knowledge for large-scale task planning,'' \emph{arXiv preprint
  arXiv:2305.14078}, 2023.

\bibitem{ocker2023commonsense}
F.~Ocker, J.~Deigm{\"o}ller, and J.~Eggert, ``Exploring {Large Language Models}
  as a source of common-sense knowledge for robots,'' in \emph{International
  Semantic Web Conference}, 2023.

\bibitem{ding2023leveraging}
Y.~Ding, X.~Zhang, C.~Paxton, and S.~Zhang, ``Leveraging commonsense knowledge
  from large language models for task and motion planning,'' \emph{RSS 2023
  Workshop on Learning for Task and Motion Planning}, 2023.

\bibitem{brown2020language}
T.~Brown, B.~Mann, N.~Ryder, M.~Subbiah, J.~D. Kaplan, P.~Dhariwal,
  A.~Neelakantan, P.~Shyam, G.~Sastry, A.~Askell \emph{et~al.}, ``Language
  models are few-shot learners,'' \emph{Advances in neural information
  processing systems}, 2020.

\bibitem{li2022pre}
S.~Li, X.~Puig, C.~Paxton, Y.~Du, C.~Wang, L.~Fan, T.~Chen, D.-A. Huang,
  E.~Aky{\"u}rek, A.~Anandkumar \emph{et~al.}, ``Pre-trained language models
  for interactive decision-making,'' \emph{Advances in Neural Information
  Processing Systems}, 2022.

\bibitem{yang2023harnessing}
J.~Yang, H.~Jin, R.~Tang, X.~Han, Q.~Feng, H.~Jiang, B.~Yin, and X.~Hu,
  ``Harnessing the power of {LLMs} in practice: A survey on {ChatGPT} and
  beyond,'' \emph{arXiv preprint arXiv:2304.13712}, 2023.

\bibitem{joublin2023glimpse}
F.~Joublin, A.~Ceravola, J.~Deigmoeller, M.~Gienger, M.~Franzius, and
  J.~Eggert, ``A glimpse in {ChatGPT} capabilities and its impact for {AI}
  research,'' \emph{arXiv preprint arXiv:2305.06087}, 2023.

\bibitem{yang2023foundation}
S.~Yang, O.~Nachum, Y.~Du, J.~Wei, P.~Abbeel, and D.~Schuurmans, ``Foundation
  models for decision making: Problems, methods, and opportunities,''
  \emph{arXiv preprint arXiv:2303.04129}, 2023.

\bibitem{kambhampati2023can}
\BIBentryALTinterwordspacing
S.~Kambhampati, ``Can {LLMs} really reason and plan?'' \emph{Blog at
  Communications of the ACM}, 2023. [Online]. Available:
  \url{https://cacm.acm.org/blogs/blog-cacm/276268-can-llms-really-reason-and-plan/fulltext}
\BIBentrySTDinterwordspacing

\bibitem{huang2022language}
W.~Huang, P.~Abbeel, D.~Pathak, and I.~Mordatch, ``Language models as zero-shot
  planners: Extracting actionable knowledge for embodied agents,''
  \emph{International Conference on Machine Learning}, 2022.

\bibitem{xie2023translating}
Y.~Xie, C.~Yu, T.~Zhu, J.~Bai, Z.~Gong, and H.~Soh, ``Translating natural
  language to planning goals with {L}arge-language models,'' \emph{arXiv
  preprint arXiv:2302.05128}, 2023.

\bibitem{singh2023progprompt}
I.~Singh, V.~Blukis, A.~Mousavian, A.~Goyal, D.~Xu, J.~Tremblay, D.~Fox,
  J.~Thomason, and A.~Garg, ``Progprompt: Generating situated robot task plans
  using {Large Language Models},'' \emph{IEEE International Conference on
  Robotics and Automation (ICRA)}, 2023.

\bibitem{wang2023describe}
Z.~Wang, S.~Cai, A.~Liu, X.~Ma, and Y.~Liang, ``Describe, explain, plan and
  select: Interactive planning with {Large Language Models} enables open-world
  multi-task agents,'' \emph{arXiv preprint arXiv:2302.01560}, 2023.

\bibitem{valmeekam2022large}
K.~Valmeekam, A.~Olmo, S.~Sreedharan, and S.~Kambhampati, ``{Large Language
  Models} still can't plan ({A Benchmark for {LLMs} on Planning and Reasoning
  about Change},'' \emph{arXiv preprint arXiv:2206.10498}, 2022.

\bibitem{liu2023llm+}
B.~Liu, Y.~Jiang, X.~Zhang, Q.~Liu, S.~Zhang, J.~Biswas, and P.~Stone,
  ``{LLM+P}: Empowering {Large Language Models} with optimal planning
  proficiency,'' \emph{arXiv preprint arXiv:2304.11477}, 2023.

\bibitem{valmeekam2023planning}
K.~Valmeekam, M.~Marquez, S.~Sreedharan, and S.~Kambhampati, ``On the planning
  abilities of {Large Language Models} -- {A} critical investigation,''
  \emph{arXiv preprint arXiv:2305.15771}, 2023.

\bibitem{guan2023leveraging}
L.~Guan, K.~Valmeekam, S.~Sreedharan, and S.~Kambhampati, ``Leveraging
  pre-trained {Large Language Models} to construct and utilize world models for
  model-based task planning,'' \emph{arXiv preprint arXiv:2305.14909}, 2023.

\bibitem{raman2022planning}
S.~S. Raman, V.~Cohen, E.~Rosen, I.~Idrees, D.~Paulius, and S.~Tellex,
  ``Planning with {Large Language Models} via corrective re-prompting,''
  \emph{arXiv preprint arXiv:2211.09935}, 2022.

\bibitem{ichter2023}
B.~Ichter, A.~Brohan, Y.~Chebotar, C.~Finn, K.~Hausman, A.~Herzog, D.~Ho
  \emph{et~al.}, ``Do as {I} can, not as {I} say: Grounding language in robotic
  affordances,'' \emph{Conference on Robot Learning}, 2023.

\bibitem{huang2022inner}
W.~Huang, F.~Xia, T.~Xiao, H.~Chan, J.~Liang, P.~Florence, A.~Zeng, J.~Tompson,
  I.~Mordatch, Y.~Chebotar \emph{et~al.}, ``{Inner Monologue}: Embodied
  reasoning through planning with language models,'' \emph{arXiv preprint
  arXiv:2207.05608}, 2022.

\bibitem{lin2023text2motion}
K.~Lin, C.~Agia, T.~Migimatsu, M.~Pavone, and J.~Bohg, ``Text2motion: From
  natural language instructions to feasible plans,'' \emph{arXiv preprint
  arXiv:2303.12153}, 2023.

\bibitem{rana2023sayplan}
K.~Rana, J.~Haviland, S.~Garg, J.~Abou-Chakra, I.~Reid, and N.~Suenderhauf,
  ``{SayPlan}: Grounding {Large Language Models} using {3D} scene graphs for
  scalable task planning,'' \emph{arXiv preprint arXiv:2307.06135}, 2023.

\bibitem{patra2021deliberative}
S.~Patra, J.~Mason, M.~Ghallab, D.~Nau, and P.~Traverso, ``Deliberative acting,
  planning and learning with hierarchical operational models,''
  \emph{Artificial Intelligence}, vol. 299, p. 103523, 2021.

\bibitem{gienger2010whole}
M.~Gienger, M.~Toussaint, and C.~Goerick, ``Whole-body motion
  planning--building blocks for intelligent systems,'' \emph{Motion Planning
  for Humanoid Robots}, p.~67, 2010.

\bibitem{mcdermott_pddl_1998}
D.~McDermott, M.~Ghallab, A.~Howe, C.~Knoblock, A.~Ram, M.~Veloso, D.~Weld, and
  D.~Wilkins, ``{PDDL} - {The} {Planning} {Domain} {Definition} {Language},''
  \emph{The AIPS-98 Planning Competition Committee}, 1998.

\bibitem{haslum2019introduction}
P.~Haslum, N.~Lipovetzky, D.~Magazzeni, C.~Muise, R.~Brachman, F.~Rossi, and
  P.~Stone, \emph{An introduction to the {Planning Domain Definition
  Language}}.\hskip 1em plus 0.5em minus 0.4em\relax Springer, 2019, vol.~13.

\bibitem{bharadhwaj2023roboagent}
H.~Bharadhwaj, J.~Vakil, M.~Sharma, A.~Gupta, S.~Tulsiani, and V.~Kumar,
  ``Roboagent: Generalization and efficiency in robot manipulation via semantic
  augmentations and action chunking,'' \emph{arXiv preprint arXiv:2309.01918},
  2023.

\bibitem{sendhoff2020cooperative}
B.~Sendhoff and H.~Wersing, ``Cooperative intelligence-a humane perspective,''
  in \emph{2020 IEEE International Conference on Human-Machine Systems
  (ICHMS)}.\hskip 1em plus 0.5em minus 0.4em\relax IEEE, 2020, pp. 1--6.

\end{thebibliography}

\end{document}